\documentclass[sigconf]{acmart}
\usepackage{multirow}
\usepackage{booktabs}
\usepackage{makecell}
\usepackage{graphicx}

\AtBeginDocument{%
  \providecommand\BibTeX{{%
    \normalfont B\kern-0.5em{\scshape i\kern-0.25em b}\kern-0.8em\TeX}}}

\setcopyright{acmcopyright}
\copyrightyear{2018}
\acmYear{2018}
\acmDOI{10.1145/1122445.1122456}

\begin{document}

\title{Interest-oriented Universal User Representation via Contrastive Learning}



\author{Qinghui Sun, Jie Gu, Bei Yang, XiaoXiao Xu, Renjun Xu, Shangde Gao, Hong Liu, Huan Xu}
\email{{yuyang.sqh,yemu.gj,bella.yb,xiaoxiao.xuxx,liuhong.liu,huan.xu}@alibaba-inc.com}
\affiliation{%
  \institution{Alibaba Group}
  \country{China}
}

\begin{abstract}
User representation is essential for providing high-quality commercial services in industry. Universal user representation has received many interests recently, with which we can be free from the cumbersome work of training a specific model for each downstream application. In this paper, we attempt to improve universal user representation from two points of views. First, a contrastive self-supervised learning paradigm is presented to guide the representation model training. It provides a unified framework that allows for long-term or short-term interest representation learning in a data-driven manner. Moreover, a novel multi-interest extraction module is presented. The module introduces an interest dictionary to capture principal interests of the given user, and then generate his/her interest-oriented representations via behavior aggregation. Experimental results demonstrate the effectiveness and applicability of the learned user representations.
\end{abstract}


\maketitle

\section{Introduction}
User modeling, an essential technique for developing personalized services (\emph{e.g.}, in recommender and advertising systems), helps business to improve user experience and create greater business value. One critical issue in user modeling is to characterize users with embeddings according to massive historical behavior data \cite{ding2017multi,andrews2019learning,gu2020exploiting}. During this process, an encoder is required to convert user behaviors into low-dimensional representation. The obtained representation is expected to contain rich information, and being capable of capturing diverse interests of users (\emph{i.e.}, being interest-aware). It plays an important role in many applications, like recommendation, user preference prediction and user profiling.

Most existing user modeling methods are elaborately designed for certain tasks. They generate task-specific user representation, which is simultaneously learned with the downstream classifier. For example,  Zhou \emph{et al.} \cite{zhou2019deep} proposes a Deep Interest Evolution Network (DIEN) for click rate prediction (CTR), which models users from their click behaviors on the e-commerce platform. Covington \emph{et al.} \cite{covington2016deep} characterizes users from their watched videos to improve the experience of video recommendation. These task-specific user representations are satisfactory for their own downstream applications. However, they can hardly be used on other unseen tasks due to the limitation of the supervised training mechanism.  The poor generalization ability limits the usage of task-specific representations in real-world applications. Actually in our business scenarios, there are dozens or even hundreds of downstream tasks. Training a particular model for each application would be prohibitively expensive. Thus, developing task-independent user representation models is of great demand. 

The universal (a.k.a., general-purpose) user representation can be seen as a compressed expression of user behaviors without specific task biases. The pre-trained universal representations can be seamlessly and directly applied to various downstream applications (without fine-tuning). We only need to further train a simple MLP for certain downstream task, instead of training the entire task-specific model. Such a pipeline shows great advantages in reducing time, manpower and material resources. 

Though learning universal user representation is important and interesting, the study of this subject is still in its early-stage. Only a few previous works focus on this challenging topic \cite{ding2017multi,andrews2019learning,gu2020exploiting}. In general, fully mining valuable user information with well-designed pretext task and encoder is the key to the success of the universal representation. This work attempts to improve the performance and applicability of user representations from these two aspects. Specifically, a contrastive self-supervised learning paradigm and a multi-interest oriented encoder are presented.

Users usually have similar behavior patterns over different time periods \cite{wang2020calendar,gu2020exploiting}. Given a user, such patterns reveals his/her interests, and thus should be captured by the representation and distinguishable from other users. With this spirit, we introduce contrastive learning to build the pretext task to guide the representation model training. During training, the encoded representations of the same person are considered as positive pairs. The rest user representations within a mini-batch are treated as negative examples. Specifically, given representations of two time periods, the representations  belonging to the same person should be more similar than the others belonging to different persons. The model learns high-quality representations by identifying and distinguishing behavior patterns of people via a contrastive loss in the latent space.

The behavior patterns of users are effected by not only long-term properties but also short-term interests. Our contrastive representation learning model is a unified framework. By changing the time span of input behaviors, the representation model is able to capture long-term or short-term interests. For example, suppose that a man has many search logs about baits or fishing equipments over a long time span, while reviews fishing rods recently. The model would be driven to mine the underlying fact that the user has a hobby of fishing when we encode representations with behaviors of a long period. In contrast, if the representations are extracted with only the recent behaviors, the model is driven to capture the information that the user has interests about fishing rods lately.

Besides the learning objective, improving the model capability of capturing diverse interests of users is another key of our work. To this end, a multi-interest oriented representation encoder is proposed. Unlike previous works \cite{pal2020pinnersage,gu2020exploiting}, we introduce a trainable dictionary to represent extensive interests explicitly. For each behavior of a given user, the most relevant interest is first indicated with the dictionary. Accordingly, his/her principal interests can be well captured and inferred. Then multiple interest-oriented user representations can be reasonably generated by aggregating corresponding behaviors. In practice, a regulation loss is additionally introduced. It benefits optimization by adaptively adjusting the parameter updates in interest dictionary.

The main contributions of this work are summarized as follows:

(1) We present interest-oriented contrastive learning as a paradigm of building effective universal user representations with large-scale unlabeled behavior data. The performance advantages of contrastive representation learning are empirically verified by extensive experiments on two real-world datasets. The learned user representation is capable of handling various downstream applications.

(2) Our model provides a unified framework for universal user representation learning. Long-term or short-term interest representations can be learned in a data-driven manner, by only changing the time span of input behaviors. An interesting insight is that the contrastive representation model can adaptively mine the underlying factors (long-term properties or short-term interests) according to the given behavior data.

(3) A multi-interest extraction module is presented to improve the model capability of capturing diverse interests of users. A trainable dictionary is introduced to represent interests explicitly, with which multiple interest-oriented representations can be reasonably generated for the given user. Moreover, a regularization loss is further introduced to reduce the variance of parameter updates in interest dictionary, which would benefit the learning.

\section{Related Work}
\textbf{Universal User Representation Learning} 
Universal user representation has an advantage in reducing feature engineering and saving resources. The earliest methods adopt dimensionality reduction or word/document embedding techniques to generate user representations \cite{yu2016user,amir2016modelling,benton2016learning,ding2017multi}. Ni \emph{et al.} \cite{ni2018perceive} proposes to train universal user representation with multi-tasks. Despite the improvements, the generalization still suffers from the requirement of the annotated training set and the lack of principles of selecting proper training tasks. Recently, Gu \emph{et al.} \cite{gu2020exploiting} propose to train effective universal user representations with a novel objective named behavioral consistency loss and multi-hop attention mechanism, who firstly generic user representations via self-supervised learning. However, behavioral consistency loss is designed as a word distribution prediction task, which is satisfied for text data. The raw data usually contains massive uninformative words, and certainly these noises would disturb the information mining. In this paper, we propose to use contrastive learning to learn user characteristics from different time periods of behaviors. The learning is directly performed at the representation level, which is more natural and reasonable.

\textbf{Contrastive Learning} 
Contrative learning is a popular method in representation learning in Computer Vision, which is proposed for grouping similar samples closer and diverse the different samples far from each other by a similarity metric. Usually the augmented version of the original image is seen as a positive sample, the rest of the samples in the batch are considered negative samples\cite{he2020momentum, chen2020simple,caron2020unsupervised} by the model. The model is supposed to learn key information of samples for distinguishing the positive samples from many negative samples. This way pushes the model to learn quality representations. Recent methods SwaV \cite{caron2020unsupervised}, MoCo \cite{he2020momentum}, SimCLR \cite{chen2020simple} have produced results comparable to the supervised method on ImageNet \cite{deng2009imagenet}. Due to the great success of contrastive learning in various computer vision tasks. In Natural Language Processing (NLP), contrastive learning also has shown significant improvement on NLP tasks such as text representations learning \cite{giorgi2020declutr}, cross-lingual pre-training \cite{chi2020infoxlm} and so on. Users usually have similar behavior patterns in a period of time \cite{wang2020calendar,gu2020exploiting}, which reflects users' interests and characteristics. It is convincing to utilize contrastive learning to discover such patterns and make them distinguishable from those of other users. To the best of our knowledge, this work is one of the pioneering works that applies contrastive learning to universal user representation modeling.

\textbf{Multi-interests User Representations}
Multi-interest user representations are first proposed by Li \emph{et al.} \cite{li2019multi}, which introduces capsule network and dynamic routing mechanism to cluster past behaviors to generate multiple interest representations for each user in recommendation areas. Cen \emph{et al.} \cite{cen2020controllable} propose a controllable multi-interests framework to balance the recommendation accuracy and diversity, which explores dynamic routing method and self-attentive methods to group user behaviors into multi-interests representations. In our view, capsule module can only extract user interests individually (by clustering each user's behaviors independently). Accordingly, the mapping from behaviors to underlying interests is not constant, which is against the common sense. Self-attentive method are limited by the attention mechanism, which is lack of interpretability and expressive ability. To solve these problems, an interest dictionary is introduced in this work, with which behaviors can be consistently oriented to unified underlying interests for any given users. It collects and infers several principal user interests representations, which benefits learning the representation learning. In addition, we introduce a  regulation loss to update each interest vector evenly.

\begin{figure*}[t]
    \centering
    \includegraphics[width=1\textwidth]{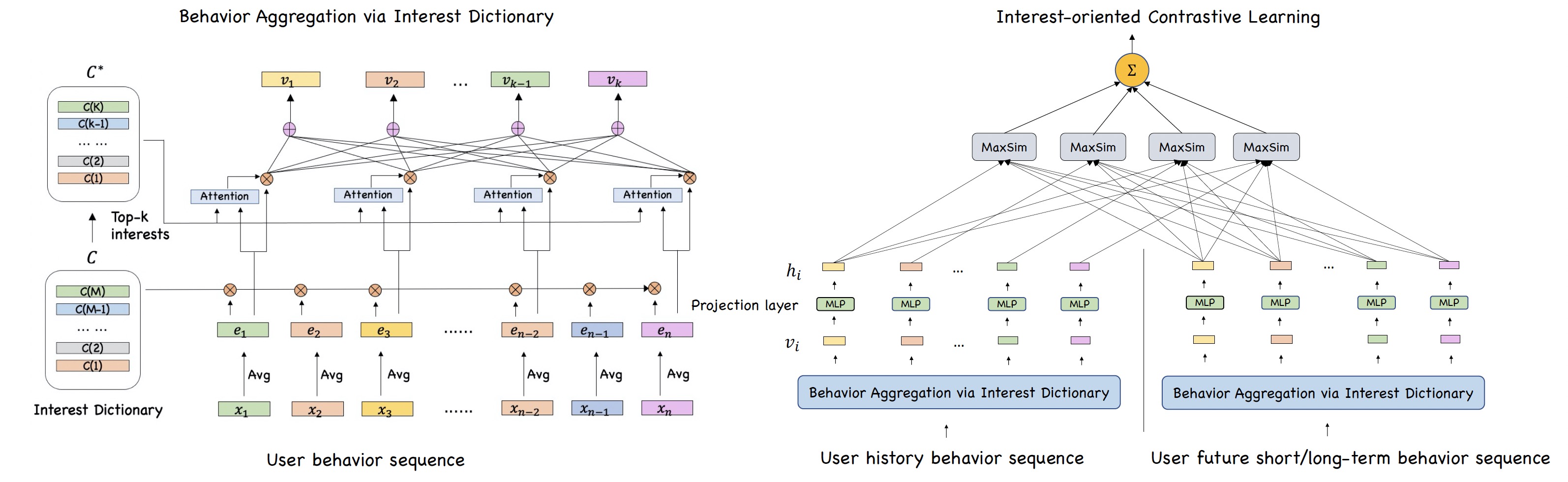}
        \caption{Illustration of our approach for learning universal user representations. The input of our model is a sequence of user behaviors. With an interest dictionary, the principal interests of the given user can be inferred. Then his/her multiple interest-oriented representations can be generated according to the chosen interests. A novel pretext task, namely interest-oriented contrastive learning, is proposed to guide the representation model training.}
    \label{figure:overview}
\end{figure*}

\section{Method}
Given a user, his/her behaviors can be formulized as a set, \emph{i.e.}, $\mathbf{S}=\{x_1, \dots, x_{|\mathbf{S}|}\}$. Specifically, $x_i$ denotes the context or content of the $i$-th behavior, and $|\mathbf{S}|$ is the total number of the historical behaviors. The goal of user representation learning is to achieve an encoder being capable of generating low-dimensional yet informative representations based on historical behaviors, \emph{i.e.},
\begin{equation}
\mathbf{V}=f_{encoder}(\mathbf{S})\label{enco},
\end{equation}
where $ \mathbf{V} = (v_1,v_2,\cdots,v_K)\in \mathbf{R}^{D*K} $ denotes the generated representation vectors of the given user. $D$ and $K$ are the dimension and number of the obtained representation vectors, respectively. It is noteworthy that here we introduce multiple representation vectors to explicitly indicate different interests of users. When $K$ equals 1, it degenerates to the case of single user representation as in previous works\cite{amir2016modelling, gu2020exploiting, ni2018perceive}.

The pipeline of this work is illustrated in Figure \ref{figure:overview}. There are two main components. The first one is a behavior aggregation module, which learns to transform the behavior data, \emph{i.e.}, $\mathbf{S}$, into multiple representation vectors as in Equation \ref{enco}. Details are given in the section 3.2. The other one is the Interest-oriented Contrastive Learning, which is a novel pretext task to guide the representation learning. This pretext task is radically different from previous ones, as the self-supervised learning is built between representation sets rather than single representations. Moreover, it is a unified learning framework, within which long-term and short-term interests can be well captured and characterized in a data-driven manner. Refer to section 3.3 for more details.

\subsection{Behavior Embedding}
Text data is ubiquitous in user logs and usually contains rich information. Without loss of generality, we focus on the text modality in this work, \emph{e.g.}, the reviewed item titles or search queries. Formally, the $i$-th behavior $x_i \in \mathbf{S}$ can be expressed as $x_i = [w_1^i, \cdots, w_{|x_i|}^i]$, where $w_j^i$ represents the $j$-th word in $x_i$, drawn from a vocabulary $O$. $|x_i|$ denotes the length of $x_i$ (number of words). We adopt average pooling to produce the behavior embedding $e_i$, \emph{i.e.},
 \begin{equation}
e_i = Avg(E(w_1^i), \cdots, E(w_{|x_i|}^i)),
 \end{equation}
 where $E(\cdot)$ indicates the table look-up operation. $Avg(\cdot)$ denotes the average pooling operation.

\subsection{Aggregation with Interest Dictionary}
 We propose a learnable dictionary-based network module to convert behavior data into multiple user representations (acting as the $f_{encoder}$ in Equation \ref{enco}. A trainable interest dictionary $\mathbf{C} = (c_1,...,c_M)$ is introduced, where $c_i$ represents a single interest vector and $M$ denotes the dictionary size (we set $M > K$). These $M$ interest vectors can be regarded as anchors, with which principal interests of users can be first inferred and then historical behaviors can be reasonably aggregated to generate the representations $\mathbf{V}$. Such a multi-interest extraction process improves the model capability of capturing diverse interests of users. Specifically, we first calculate the relevance scores between each interest vector and behavior embedding. The score $p_{ij}$ between $c_i$ (1 $ \leq i\leq M$) and $e_j$ (1 $ \leq j \leq|\mathbf{S}|$) can be computed as
\begin{equation}
 p_{ij}=\frac{Dot(c_i,e_j)}{||c_i||\cdot||e_j||},
\end{equation}
 where $Dot(a,b)$ denotes the dot product of the vectors $a$ and $b$ and $||a||$ is the l2-normalization of $a$. The accumulated relevance score of the interest vector $c_i$ over all behavior embeddings of the given user can be computed as 
\begin{equation}
 P_i = \sum_{j=1}^{|\mathbf{S}|}p_{ij} (1 \leq i \leq M)\label{sum_p}.
\end{equation}
A large $P_i$ indicates that there are many behaviors being relevant to the interest vector $c_i$. That is, the larger the $P_i$ is, the more likely the given user has the corresponding interest. Thus we can obtain the principal interests of the given user by locating Top-$K$ indices of $P$, \emph{i.e.},
\begin{equation}
\{d_1,...,d_K\} = Indice_{K}(P_1,...,P_i,...P_M),
\end{equation}
where $Indice_{K}(\cdot)$ denotes the operation of locating Top-$K$ indices with the value of $P_i$. Accordingly, a set of interest vectors can be collected with the indices $\{d_1,...,d_K\}$ from the interest dictionary $\mathbf{C}$, explicitly indicating the Top-$K$ principal interests: 
\begin{equation}
\mathbf{C}^* = Index(\{c_1,...,c_M\},\{d_1,...,d_K\}),
\end{equation}
where $\mathbf{C}^* = (c^{*}_1,...,c^{*}_K)$ represents the collected Top-$K$ interest vectors. $Index(\mathbf{A},\mathbf{B})$ denotes the operation of selecting corresponding values from set $\mathbf{A}$ according to the indices in set $\mathbf{B}$. 

After the principal interests inference, an attention module is further utilized to generate multi-interest user representations. For each interest vector in $\mathbf{C}^*$  and each behavior embedding $e_j$, the attention weight $\alpha_{ij}$ is computed as
\begin{equation}
\alpha_{ij}=\frac{\exp(Dot(c^{*}_i,e_j))}{\sum_{k=1}^{|S|}\exp(Dot(c^{*}_i,e_k))}.
\end{equation}
Then $K$ interest-oriented user representations can be obtained by aggregating behavior embeddings with attention weights. The representation corresponding to the interest vector $c_i^{*}$ is computed as 
\begin{equation}
 v_i = \sum_{j=1}^{|S|}\alpha_{ij}e_j (1 \leq i \leq K).
\end{equation}

During training, we notice that only a few interest vectors are frequently updated, while others in $\mathbf{C}$ are not. To address this issue, we introduce a regularizer to adaptively adjust the parameter update frequencies of interest vectors to improve training. The regularizer can be expressed as 
\begin{equation}
\hat{P_i}=\frac{1}{B}\sum_{b=1}^{B}P_i^{b} (1 \leq b \leq B)
\end{equation}
\begin{equation}
L_{reg}=\sum_{i=1}^{M}(\hat{P_i}-\frac{1}{M}\sum_{i=1}^{M}{\hat{P_i}})^{2},
\end{equation}
where $B$ is the batch size, $P_i^b$ denotes the $P_i$ (computed by Equation \ref{sum_p}) of the $b$-th training sample in a batch. $L_{reg}$ penalizes the case that behaviors are always associated with several certain interest vectors (only these frequently selected interest vectors would fully participate in training). It helps to balance the utilization of interest vectors. Then the entire interest dictionary would be sufficiently trained.

 \begin{figure}[t]
    \centering
    \includegraphics[width=0.5\textwidth]{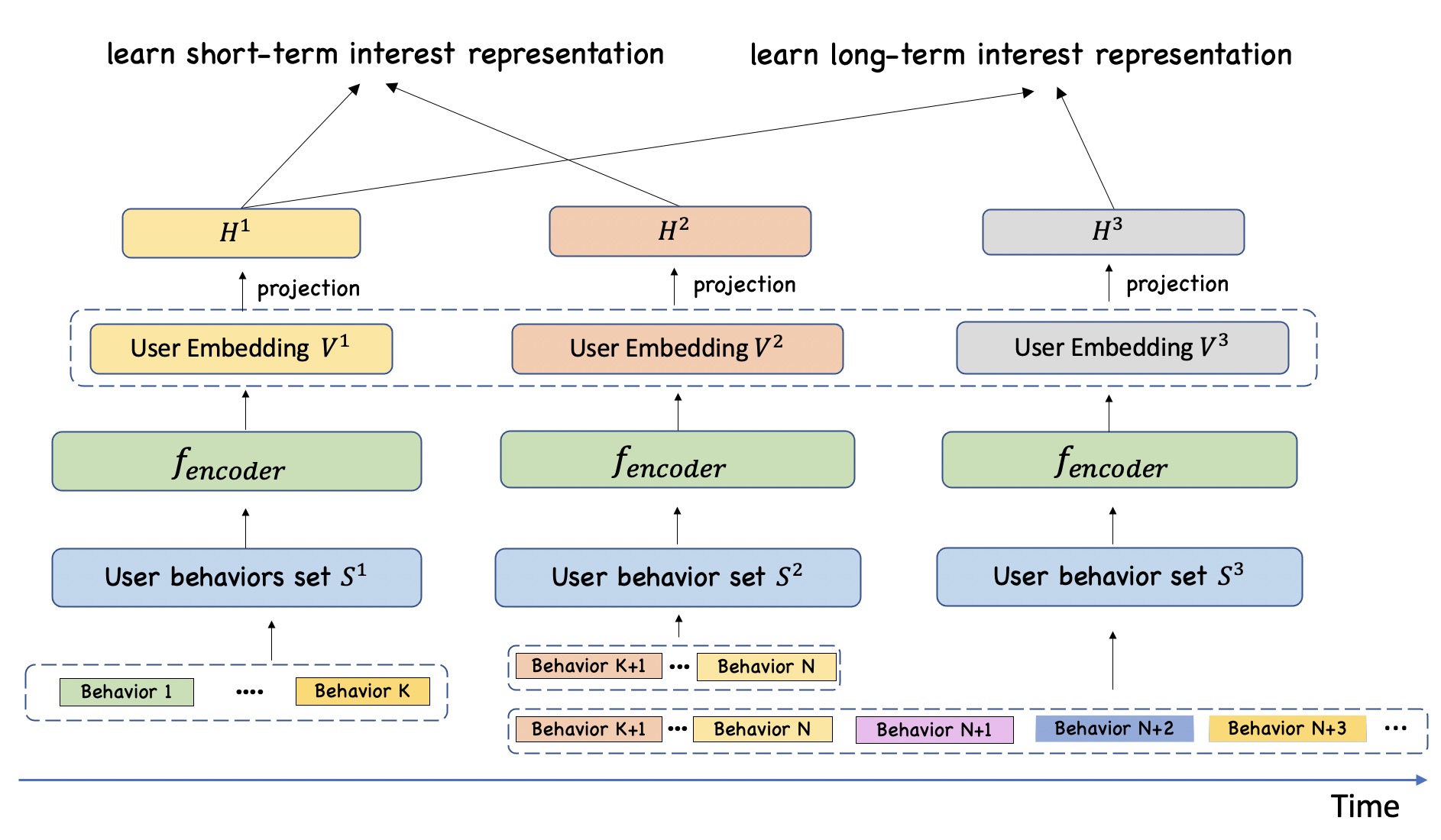}
        \caption{Illustration of splitting user behaviors into three sets for short-term interest or long-term interest user representation learning}
    \label{figure:pretext}
\end{figure}

\subsection{Interest-oriented Contrastive Learning}
User behaviors are usually correlated, effected by some latent factors like user property and interests. It is natural and reasonable to introduce contrastive learning to characterize such latent dependencies between behaviors. In this work, a new framework is presented for contrastive multi-interest user representation learning. The learning is performed by maximizing agreements between two groups of interest-oriented representations of the same user (generated with behaviors collected from two time periods) via a contrastive loss. Moreover, it is a unified framework for producing short-term or long-term user representations. By changing the time span of input behaviors, the model would be driven to mine correlations behind behaviors casued by short-term or long-term interests of users.

The training is conducted on two behavior sets, \emph{i.e.}, a historical set $\mathbf{S}^1$ and a target set. The target set consists of behaviors that are produced after historical ones in terms of time. In practice, we collect two types of target sets, denoted as $\mathbf{S}^2$ and $\mathbf{S}^3$, for learning short-term and long-term user representations respectively. As shown in Figure \ref{figure:pretext},  $\mathbf{S}^2$ consists of user behaviors in the short future and $\mathbf{S}^3$ contains user behaviors in a long period of time. For each set, behaviors are aggregated via interest dictionary to generate multi-interest user representations, \emph{i.e.},
\begin{equation}
\mathbf{V}^i = f_{encoder}(\mathbf{S}^i)
\end{equation}
where $\mathbf{V}^i=(v^i_1, v^i_2, \cdots, v^i_K) \in \mathbf{R}^{D*K}, i=(1,2,3)$.

\textbf{Projection Head} 
Hinton \emph{et al.} have shown that using a projection head (a nonlinear transformation) substantially benefits the quality of the learned representation \cite{chen2020simple}. It is usually implemented by a small neural network that maps representations to a latent space where the pretext task is applied. Formally, the mapping can be expressed as 
\begin{equation}
\mathbf{H}^i= g(\cdot) = W_{(2)}\sigma(W_{(1)}(\mathbf{V}^i)),
\end{equation}
where $g(.)$ represents the projector (an MLP in this work), $W_{(1)}$ and $W_{(2)}$ are the parameters, $\sigma(\cdot)$ is a non-linearity activation function, and $\mathbf{H}^i = (h^i_1, h^i_2,\cdots,h^i_K), (i=1,2,3)$. We find it beneficial to define the pretext task on $\mathbf{H}^i$ rather than $\mathbf{V}^i$. 

Long-term and short-term user representations are learned separately, by using the same contrastive loss while different behavior sets. The representation model learns to extract and characterize short-term interests of users when the contrastive leaning is build on the pairs of embeddings in $\mathbf{H}^1$ and $\mathbf{H}^2$. Instead, the model focus on long-term interests when using $\mathbf{H}^1$ and $\mathbf{H}^3$.

\textbf{Short-term Interest Extraction} 
If a user has some short-term interest, he/she would pay close attention to associated products in a relatively short time. For example, if a man wants to buy some electronic product for watching videos, he would probably have many search logs about computers or pads over a short period of time. Thus, there should exist correlations between behaviors in $\mathbf{S}^1$ and $\mathbf{S}^2$, and the representations $\mathbf{H}^1$ and $\mathbf{H}^2$ should have some kind of "agreement".

We randomly sample a minibatch of $B$ users. The representation sets $\mathbf{H}^1$ and $\mathbf{H}^2$ derived from the same user are treated as a positive pair. Similar to \cite{chen2020simple,2020Momentum}, $\mathbf{H}^1$ and $\mathbf{H}^2$ derived from different users within the minibatch are treated as negative pairs. For convenience, we denote negative pairs as $(\mathbf{H}^1, \mathbf{H}^2_b), (b=1,2,\ldots,B)$ and $\mathbf{H}^2_b \neq \mathbf{H}^2$. The contrastive loss for a positive pair is
\begin{equation}
l=-log\frac{exp(sim(\mathbf{H}^1,\mathbf{H}^2)/\tau)}{\sum_{b}exp(sim(\mathbf{H}^1,\mathbf{H}_b^2)/\tau)},
\label{sim_h}
\end{equation}
where $\tau$ denotes a temperature parameter, $sim(\mathbf{H}^1,\mathbf{H}^2)$ measures the relevance of the two representation sets $\mathbf{H}^1$ and $\mathbf{H}^2$, which will be introduced in the following. With the pretext task, the representation model learns to identify $\mathbf{H}^2$ from $\mathbf{H}^2_b, b=(1,2,\ldots,B)$ based on $\mathbf{H}^1$. The dependencies between behaviors caused by short-term interests would be well mined and characterized.

Since we use multiple embeddings to represent users, how to measure the "agreement" of $\mathbf{H}^1$ and $\mathbf{H}^2$ remains to be solved in Equation \ref{sim_h}. An intuitive idea is to concatenate all $h^i_1,h^i_2,\cdots,h^i_K (i=1,2)$ and then calculate the distance. However, this may break the similarity relationships between representations, and thus hurt the performance. Here we propose a new way. Specifically, for each interest-oriented user representation pair $h^1_i \in \mathbf{H}^1$ and $h^2_j \in \mathbf{H}^2$, we compute their cosine distances. According to these distances, for each $h^1_i \in H1$, we locate its most relevant pair $(h^1_i,h^2_j \in \mathbf{H}^2)$. Then all distance values of the chosen pairs could be summed as a measure of the relevance of the two representation sets. Formally, the process can be expressed as
\begin{equation}
sim(\mathbf{H}^1,\mathbf{H}^2) = \sum_{i\in|\mathbf{H}^1|}{max}_{j\in|\mathbf{H}^2|}\frac{Dot(h_{i}^1,h_{j}^2)}{||h_{i}^1||\cdot||h_{j}^2||},
\end{equation}
where $h_{i}^1$ denotes the $i$-th interest-oriented representation in $\mathbf{H}^1$. $h_{j}^2$ is the $j$-th interest-oriented representation in $\mathbf{H}^2$. $|\mathbf{H}^1|$ and $|\mathbf{H}^2|$ denotes the size of $\mathbf{H}^1$ and $\mathbf{H}^2$.

The final objective is defined as the sum of contrastive losses of all positive pairs in a mini-batch, \emph{i.e.},
\begin{equation}
L_{short} = {\beta}\sum_{i=1}^{B}l_i + {\gamma}L_{reg}
\label{loss_short},
\end{equation}

where $\beta$ and $\gamma$ are hyper-parameters. There are $B$ positive pairs in equation \ref{loss_short}. 

\textbf{Long-term Interest Extraction} User behaviors are not only influenced by short-term interests but also long-term interests. There also exists consistent patterns between behaviors in a long period of time, reflected by properties, habits, preferences, etc. For instance, a man has a hobby of fishing would always search or browse products like baits and fishing rods. To enable the model to characterize such long-term patterns, behaviors over a long time span should be included in the training. Thus, the learning is performed on the behavior sets $\mathbf{S}^1$ and $\mathbf{S}^3$. Similar to the case in "short-term interest extraction", $\mathbf{H}^1$ and $\mathbf{H}^3$ derived from the same user form a positive pair, while those from difference users in a minibatch form negative ones. The relevance of representation sets can then be expressed as
\footnote{One can see that the short-term and long-term interest representations are learned within the same interest-oriented contrastive learning framework, while differing in the utilization of the time span of behaviors.}
\begin{equation}
sim(\mathbf{H}^1,\mathbf{H}^3) = \sum_{i\in|\mathbf{H}^1|}{max}_{j\in|\mathbf{H}^3|}\frac{Dot(h_{i}^1,h_{j}^3)}{||h_{i}^1||\cdot||h_{j}^3||}. 
\end{equation}
And the contrastive loss is defined as
\begin{equation}
l_i^{*}=-log\frac{exp(sim(\mathbf{H}^1,\mathbf{H}^3)/\tau)}{\sum_{b}exp(sim(\mathbf{H}^1,\mathbf{H}_b^3)/\tau)}.
\end{equation}

The overall objective for learning long-term representations of users is:
\begin{equation}
L_{long} = {\beta}\sum_{i=1}^{B}l_i^{*} + {\gamma}L_{reg}
\label{loss_long},
\end{equation}
where the loss weights $\beta$ and ${\gamma}$ are hyper-parameters.


\section{Experiments}

\begin{table}
\centering
\begin{tabular}{ccccc}
\hline
Dataset  & $|U|$ & $|\mathbf{S}|$ & $|x|$ & $|O|$ \\
\hline
Amazon & {1,725,907} & {25} & {35} & {50,000}  \\
Industry & {64,000,000} & {320} & {8} & {178,422}     \\
\hline
\end{tabular}
\caption{Statistics of the datasets.$|U|$ denotes the number of user on the dataset. $|S|$ is the number of behaviors of each user. The word length for each behavior is $|x|$. And the vocabulary size is denoted by $|O|$.}
\label{tab:data_stat}
\end{table}

\begin{table}
\newcommand{\tabincell}[2]{\begin{tabular}{@{}#1@{}}#2\end{tabular}}
\centering
\begin{tabular}{c|c|c|c}
\hline
   &Time interval & Amazon & Industry \\ \hline
\multirow{3}*{\tabincell{c}{Model \\ Pretrain}} & History $|\mathbf{S}^1|$ & 201407-201412 & 201902-201903  \\ \cline{2-4}
 &Behaviors $|\mathbf{S}^2|$ & 201501-201503 & 201904\\ \cline{2-4}
 &Behaviors $|\mathbf{S}^3|$ & 201501-201506 & 201906-201907\\ \hline
\multirow{3}*{\tabincell{c}{Tasks}} &\tabincell{c}{Model Infer}  & 201507-201512 & 201905-201906\\ \cline{2-4}
 &Short-term label & 201601-201603 & 201907 \\ \cline{2-4}
 &Long-term label & 201607-201612 & 201908-201909\\ \hline
\end{tabular}
\caption{Details of data collection. For example on Amazon dateset, user behavior data from 201407 to 201506 are split into three sets for user representation model training. Behavior data between 201507 and 201512 are used to infer user representations for downstream tasks. Behavior data from 201601 to 201603 and 201607 to 201612 are collected for labeling. positive samples for downstream evaluations.}
\label{tab:time_interval}
\end{table}

The experiments are conducted on ten downstream tasks of three types in two real-world e-commerce datasets. Each dataset consists of large-scale unlabelled user behaviors for learning universal user representations and several annotated datasets for downstream tasks. Refer to Tables \ref{tab:data_stat} and \ref{tab:time_interval} for the statistics of the datesets and the details of the data collection.

\begin{table*}
  \centering
  \label{dataset:hitrate}
  \begin{tabular}{c|c|c|ccccccc}
    \hline
     Datasets&Task&Task type&TextCNN &TF-IDF& Word2Vec & Doc2Vec & PTUM & SUMN  & Ours-ID-ICL\\  \hline
     \multirow{4}*{Amazon} 
     
     &Baseball Caps& Short-term & 0.7370  &0.6546& 0.6615 & 0.7017 & 0.7462 & 0.7561 & \textbf{0.7650}  \\ \cline{2-10}
     
     &Keyboards& Short-term & 0.8208 &0.7021& 0.7297& 0.7584 & 0.8315 & 0.8373 & \textbf{0.8481}  \\ \cline{2-10}
     
     &Video Games& Long-term & 0.7953 & 0.7360 &0.7805 & 0.7917 & 0.7933 & 0.8245  & \textbf{0.8374} \\ \cline{2-10} 
     
     &Computer& Long-term & 0.7768 &0.6733 & 0.7526 & 0.7658 & 0.7688 & 0.7948 & \textbf{0.8083}  \\ \hline
    
  \end{tabular}

\end{table*}

\begin{table*}
  \centering
  \label{dataset:hitrate}
  \begin{tabular}{c|c|c|cccccc}
    \hline
     Datasets&Task&Task type&TextCNN &TF-IDF& Word2Vec & Doc2Vec & SUMN  & Ours-ID-ICL\\  \hline

     \multirow{6}*{Industry}
     
     &Wallets&Short-term  & 0.6832&0.6650 & 0.6810 & 0.6954 & 0.7144 &\textbf{0.7388} \\ \cline{2-9}
     &Watch &Short-term  & 0.7143 &0.6864& 0.7032 & 0.7088 & 0.7048 & \textbf{0.7300}  \\ \cline{2-9}
     &Outdoor Products &Long-term  & 0.7627 & 0.6720& 0.7032 & 0.7552 & 0.7784 & \textbf{0.7896}  \\ \cline{2-9}
     &Car Accessories &Long-term  & 0.8521 &0.8118 & 0.8203 & 0.8443 &  0.8701 & \textbf{0.8781}  \\ \cline{2-9}
    &Age&User profiling &0.6064 & 0.6122& 0.5752 & 0.5351 &0.6447 & \textbf{0.6571} \\ \cline{2-9}
    &Baby Age& User profiling &0.7203 & 0.6906& 0.6666 & 0.6586 &0.7217&  \textbf{0.7387}  \\ \hline
    
  \end{tabular}
  \caption{Performance comparison on different task on the Amazon and Industry datasets. The metrics for long-term preference identification task and short-term preference identification task is AUC. The measures for user profiling task is accuracy. We use \textbf{bold} font to highlight wins.}
\label{tab:ten_task_results}
\end{table*}

\subsection{Datasets \& Implementation Details}

\textbf{Amazon Dataset\footnote{https://nijianmo.github.io/amazon/index.html}}
This dataset includes product reviews and involves product metadata like titles and categories in Amazon \cite{he2016ups}. For each user, the reviewed product titles make up a review behavior sequence. As shown in Table \ref{tab:time_interval}, for training a representation model, we selected the review logs between 2014-07 and 2014-12 to form the historical behavior set $\mathbf{S^1}$. The behavior set for short-term representation learning $\mathbf{S^2}$ contains the review logs from 2015-01 to 2015-03. The long-term behavior set $\mathbf{S^3}$ contains the review logs from 2015-01 to 2015-06.

\textbf{Industry Dataset\footnote{https://www.taobao.com/}} This dataset is built by collecting search logs on a popular e-commerce platform TaoBao in real-world scenarios as in \cite{gu2020exploiting}. The user search queries between 2019-02 and 2019-03 are collected to form the history behavior set $\mathbf{S^1}$. The short-term behavior set $\mathbf{S^2}$ consists of the search logs between 2019-04 and 2019-05. The search logs between 2019-06 and 2019-07 are collected to form the long-term behavior set  $\mathbf{S^3}$.

\textbf{Data Prepossessing}
For English texts, we perform the operations of lowercasing and word stemming. The Chinese texts are segmented by using Jieba \footnote{https://github.com/fxsjy/jieba}. A dedicated vocabulary is constructed for each dataset. We also set truncation thresholds to limit the number of behaviors $|\mathbf{S}|$, as well as the number of words $|x|$. The exceeded behaviors and words are removed. Refer to Table \ref{tab:data_stat} for all configurations.

\textbf{Parameter Configuration}
We set $M$ = 100 and $K$ = 5 for Industry dataset and $M$ = 20 and $K$ = 2 for Amazon dataset. The dimension of user representations is 256. The loss weight $\gamma$, $\beta$ are set to be 1. The loss function is optimized by Adam optimizer with a learning rate of 0.001. The batch size is set as 256. The training is stopped when loss converges on the validation set.

\subsection{Evaluations on Downstream Tasks}
Our model can generate user representations which can capture long-term or short-term interests. We adopt three types of downstream tasks to evaluate the performance. The first two types of tasks are the long-term preference identification and user profiling. They are presented to verify the performance of long-term interest representations as in \cite{gu2020exploiting}. The last one is the short-term preference identification, which is designed for the evaluation of short-term interest representations.

\begin{figure*}[t]
    \centering
    \includegraphics[width=1.0\textwidth]{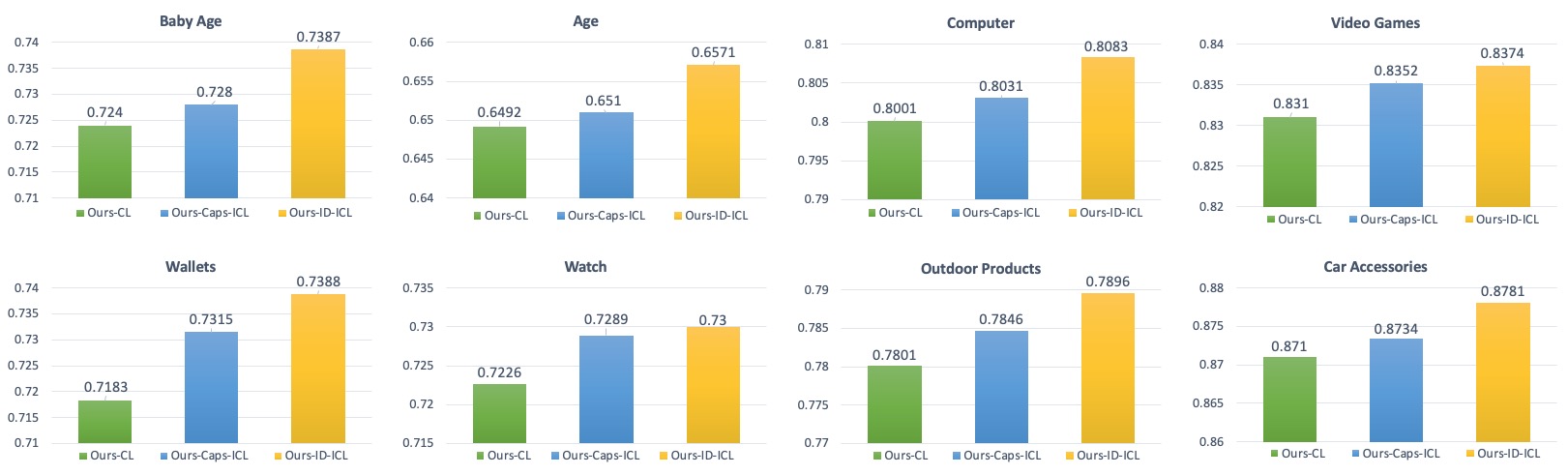}
        \caption{Ablation Study. Performance comparison of different user behavior aggregation methods. \textit{Ours-CL} utilizes max-pooling, \textit{Ours-Caps-ICL} adopts dynamic routing method. \textit{Ours-ID-ICL} uses Interest Dictionary Aggregation Module}
    \label{figure:Ablation_8}
\end{figure*}

\textbf{Task Introduction \& Data Collection}
Short-term preference identification refers to the task of predicting whether a user 
would have behaviors on the items of some category in the short future. We conduct this experiment on the Amazon and Industry datasets. For the Amazon dataset, two categories are included: Baseball Caps and Keyboards. Review logs from 2015-07 to 2015-12 are collected to infer the user representations, and a user is labeled as positive if there exists at least one review log in the target category between 2016-01 and 2016-03. For baseball caps, there are 18280 samples for training and 4636 samples for testing. For keyboards, there are 17679 samples for training and 4490 samples for testing. For the Industry dataset, two categories are considered for evaluation, including wallets and watch. We collect search queries from 2019-05 to 2019-06 for user representation inference, and transaction logs in 2019-07 for user labeling. A total of 4 million samples are collected for each category. 80\% of the samples are randomly selected for training downstream models and the rest 20\% for testing.

Long-term preference identification refers to the task of predicting whether a user would have behaviors on the commodities of a target category in the relatively long future. We follow the same data collection strategy as in \cite{gu2020exploiting} for a fair comparison. For the Amazon dataset, two categories are included: books of computers and video games. Review logs from 2015-07 to 2015-12 are collected to infer the user representations, and a user is labeled as positive if there exists at least one review log in that category between 2016-07 and 2016-12. A total of 191,856 samples are collected. For the Industry dataset, two categories including outdoor products and car accessories are considered for evaluation. Search queries from 2019-05 to 2019-06 are used for user representations inference, and the transaction logs between 2019-08 and 2019-09 are used for user labeling, which makes up a dataset containing 2.4 million samples. For all evaluation datasets, we randomly select 80\% of the samples for training downstream models and the rest for the performance testing.

User profiling prediction aims to identify user aspects such as age. We also follow the experimental configurations in \cite{gu2020exploiting}, which evaluates performance on two sub-tasks: (1) user age classification task (6-class), which predicts the age ranges of users. There are 1,628,958 samples for training and 543,561 samples for testing; (2) baby age classification task (7-class), which predicts the age ranges of users’ babies. The sizes of the training and testing sets are 396,749 and 99,411 respectively. Search queries are collected for user representation inference. The ground-truth age label comes from an anonymous questionnaire.

\textbf{Competitors}
Two types of previous representative approaches are selected for comparison. The first class of methods generates user representations without access to the annotated data labels in downstream tasks (unsupervised). The competitors include: \\
(1) \textit{TF-IDF} \cite{robertson2004understanding}
, which views texts in one's behaviors as a single document and uses a sparse statistical vector for representation; (2) \textit{Word2Vec} \cite{mikolov2013distributed}
, training word embeddings on an unlabelled corpus and computing user representations through the average of the word embeddings in behaviors; (3) \textit{Doc2Vec} \cite{le2014distributed}, regarding the behaviors of a person as a document and learning a document embedding for user representation. (4)\textit{PTUM} \cite{2020PTUM}, designed based on Bert and proposes two self-supervision tasks for user modeling pre-training.
(5) \textit{SUMN} \cite{gu2020exploiting}, training universal user representations with a novel objective named behavioral consistency loss and multi-hop attention mechanism, which firstly generate long-term user representations via self-supervised learning. The other type of competitors simultaneously learns task-specific representation encoders and classifiers in a supervised manner on downstream tasks:  \textit{TextCNN} \cite{kim2014convolutional}, applying convolution operations on the embedding concatenation of all words that appeared in behaviors and using max-pooling to get user representations.

Our method learning user representation with Interests Dictionary (ID) and Interest-oriented Contrastive Learning (ICL) modules, so we named our method as \textit{Ours-ID-ICL}

\textbf{Downstream Model}
Once the user representation is obtained, we only need a simple model for downstream predictions. The downstream model can just be an MLP classifier. The MLP has one hidden layer with dimensions of 128. The hyper-parameters of the supervised competitors are tuned on the validation set. For both the MLP and supervised models, we use Adam with a learning rate of 0.001 as the optimizer, and the batch size is set as 256.


\textbf{Results}
Table \ref{tab:ten_task_results} lists the performance comparisons on ten downstream tasks of two datasets, from which we have several observations. First, it can be seen that our method \textit{Ours-ID-ICL} improves the performance significantly against all the competitors. \textit{PTUM} and \textit{SUMN} are state-of-the-art universal user representation methods, \textit{Ours-ID-ICL} achieves over 1-3\% AUC improvements on the preference identification tasks and about 1.5\% accuracy improvement on the user profiling task. Secondly, the performance of our self-supervised method is consistently better than  the supervised-learned method \textit{TextCNN}, about 2-4\% promotions in AUC. It is worth mentioning that actually the aim of this work in not to improve universal representations to beat effective supervised competitors on certain tasks. Our user representations are shared across various downstream tasks on a dataset. Such a process has advantages in reducing feature engineering and saving resources compared to training numbers of supervised method (\emph{e.g.}, \textit{TextCNN}) for every downstream task.





\subsection{Ablation Study}
\begin{figure}[t]
    \centering
    \includegraphics[width=0.5\textwidth]{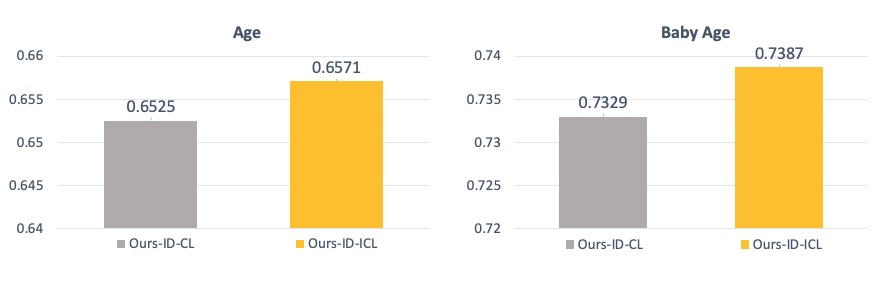}
        \caption{Ablation Study. Performance comparison of different methods with original constrastive learning and our interest-oriented constrastive learning.}
    \label{figure:Age_Baby}
\end{figure}

As shown in Figure \ref{figure:Ablation_8} and Figure \ref{figure:Age_Baby}, we conduct several ablation studies on several downstream tasks to verify the improvements of the
proposed behaviors aggregation module and the pretext task. We compare \textit{Ours-ID-ICL} with three variants. The baseline leverages max-pooling to aggregate user behaviors and original Contrastive Learning (CL) \cite{chen2020simple} for training, denoted as \textit{Ours-CL}. To show the effectiveness of our Interest Dictionary Aggregation Module (ID), we replace the multi-interests extraction module in \textit{Ours-ID-ICL} with dynamic routing method\cite{li2019multi}, denoted as \textit{Ours-Caps-ICL}. Moreover, to demonstrate the effectiveness of Interest-oriented Contrastive Learning (ICL), we concat all interest-oriented representations $h_i \in \mathbf{H}$ and then adopt original contrastive learning for training, denoted as (\textit{Ours-ID-CL}). All parameter settings and training configurations are set to be the same for all variants for a fair comparison. From Figures \ref{figure:Ablation_8} and \ref{figure:Age_Baby}, one can observe that \textit{Ours-ID-ICL} always achieves the best results, which shows that the performance improvements by Interest Dictionary Aggregation Module (ID) and Interest-oriented Contrastive Learning (ICL) are promising and stable.

\subsection{Discussion}
\begin{figure}[t]
    \centering
    \includegraphics[width=0.5\textwidth]{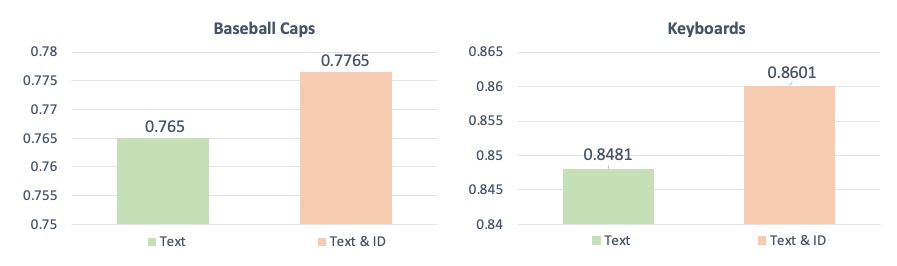}
        \caption{Performance comparison of our universal user representations with only the text data and both the text \& structured data.}
    \label{figure:Text_TextID}
\end{figure}

Our method can be seamlessly applied to deal with structured data (such as category ID, shop ID, etc.), while previous works (\emph{e.g.}, \textit{SUMN}\cite{gu2020exploiting} and \textit{PTUM}\cite{2020PTUM}) cannot. For example, one way to modify \textit{SUMN} for handling structured data is to substitute the occurrence numbers of words with those of behavior IDs. However, the size of behavior ids (such as item ID) are huge, and thus the above modification is impractical in real-worlds. Accordingly, the application ranges of \textit{SUMN} and \textit{PTUM} are limited to some extent. In contrast, our \textit{Ours-Caps-ICL} is not limited to text modalities, and the performance can be further improved by incorporating user representations generated from structured data. In this part, we measure the benefits of incorporating such structured information on two tasks in Amazon dataset. The results are shown in Figure \ref{figure:Text_TextID}. It can be seen that the AUC increases 1\% when we concatenate text-based and ID-based user representations. The generality for structured data is definitely another advantage of our method.


\begin{table}
\newcommand{\tabincell}[2]{\begin{tabular}{@{}#1@{}}#2\end{tabular}}
  \centering
  \label{dataset:hitrate}
  \begin{tabular}{p{2.3cm}p{1cm}<{\centering}p{1.1cm}<{\centering}p{1.2cm}<{\centering}}
    \hline
      Metric@ACC   &$M$=1 & $M$=50 & $M$=100 \\ \hline
     (Ours)$K$=1  &0.7240&0.7298 & 0.7268\\
     (Ours)$K$=3  &-   &0.7324 & 0.7305 \\
     (Ours)$K$=5  &-   &0.7335 & \textbf{0.7387} \\
     (Ours)$K$=7  &-   &0.7310 & 0.7312 \\
    \hline
  \end{tabular}
  \caption{Model performance of Industry Dataset for different $K$ and $M$ in Baby Age classification task.}
  \label{tab:ps1}
\end{table}

\begin{table}
\newcommand{\tabincell}[2]{\begin{tabular}{@{}#1@{}}#2\end{tabular}}
  \centering
  \label{dataset:hitrate}
  \begin{tabular}{p{2.3cm}p{1cm}<{\centering}p{1.1cm}<{\centering}p{1.2cm}<{\centering}}
    \hline
      Metric@AUC   &$M$=1 & $M$=20 & $M$=50 \\ \hline
    (Ours) $K$=1  &0.8307 &0.8310 & 0.8312\\
    (Ours) $K$=2  &-   &\textbf{0.8374} & 0.8363 \\
    (Ours) $K$=3  &-   &0.8323   & 0.8368 \\
    \hline
  \end{tabular}
  \caption{Model performance of Amazon dataset for different $K$ and $M$ in Video Games preference identification.}
  \label{tab:ps2}
\end{table}

\subsection{Hyperparameter Sensitivity}
In our model, the size of interests dictionary $M$ and the number of interests for each user $K$ are hyperparameters. We investigate the sensitivity of $K$ and $M$ of our framework. The number of behaviors per person $|\mathbf{S}|$ in the Amazon dataset is truncated to 25, and for Industry dataset is truncated to 320, so we set M to 1,20,50 and K to 1,2,3 in the Amazon dataset, then set M to 1,50,100 and K to 1,3,5 in the Industry dataset. Then we explored the effect of different combinations of M and K on the results. Table \ref{tab:ps1} and \ref{tab:ps2} illustrate the performance of our framework when the hyperparameters $K$ and $M$ change. From the results, firstly we found that keeping M unchanged, appropriately increasing K can improve the performance of the model. But when K is too large, the performance of the model will be impaired. Our analysis is due to the redundancy of the representation and the inclusion of noise. Secondly, the result shows different properties of these hyperparameters when in different datasets. For industry dataset, we find that $K = 5$ and $M = 100$ obtains the best performance, while $K = 2$ and $M = 20$ obtains superior performance on Amazon dataset. We think the reason of this phenomenon is that there are more behaviors for each user on Industry dataset than the Amazon dataset.

\section{Conclusion}
In this paper, we propose a novel learning objective named Interest-oriented Contrastive Learning as a unified framework to extract user short-term and long-term interests user representation through large-scale unlabeled user behaviors data. To capture diverse interests of users,  we introduce an Interest Dictionary Aggregation Module and a  regularization loss to aggregate user behaviors. We conduct experiments on three types of tasks in two real-world datasets to show that the proposed methods outperforms state-of-the-art user representation learning methods.

\bibliographystyle{ACM-Reference-Format}
\bibliography{anthology}



\end{document}